# Sticking to the Mean: Detecting Sticky Tokens in Text Embedding Models

Kexin Chen[1]    Dongxia Wang[1,3][†]    Yi Liu[2][†]    Haonan Zhang[1]    Wenhai Wang[1]
[1]Zhejiang University    [2]Quantstamp
[3]Huzhou Institute of Industrial Control Technology
{kxchen, dxwang, haonanzhang, zdzzlab}@zju.edu.cn,   yi009@e.ntu.edu.sg

## Abstract

Despite the widespread use of Transformer-based text embedding models in NLP tasks, surprising "sticky tokens" can undermine the reliability of embeddings. These tokens, when repeatedly inserted into sentences, pull sentence similarity toward a certain value, disrupting the normal distribution of embedding similarities and degrading downstream performance. In this paper, we systematically investigate such anomalous tokens, formally defining them and introducing an efficient detection method, **S**ticky **T**oken **D**etector (**STD**), based on sentence and token filtering. Applying STD to 40 checkpoints across 14 model families, we discover a total of 868 sticky tokens. Our analysis reveals that these tokens often originate from special or unused entries in the vocabulary, as well as fragmented subwords from multilingual corpora. Notably, their presence does not strictly correlate with model size or vocabulary size. We further evaluate how sticky tokens affect downstream tasks like clustering and retrieval, observing substantial performance degradation that approaches 50% in certain cases. Through attention-layer analysis, we show that sticky tokens disproportionately dominate the model's internal representations, raising concerns about tokenization robustness. Our findings show the need for better tokenization strategies and model design to mitigate the impact of sticky tokens in future text embedding applications.

 https://github.com/March-7/StickyToken

## 1 Introduction

Dense vector representations of text, often called text embeddings, capture semantic content and power a wide range of downstream applications, such as retrieval, classification, clustering, and semantic similarity tasks (Mikolov et al., 2013; Devlin et al., 2018; Muennighoff et al., 2023). The embedding-based retriever is also a critical component for retrieval-augmented generation (RAG) (Gao et al., 2024), which allows large language model (LLM) to access the most up-to-date external or proprietary knowledge without modifying parameters (Lewis et al., 2021).

[†]Corresponding Author

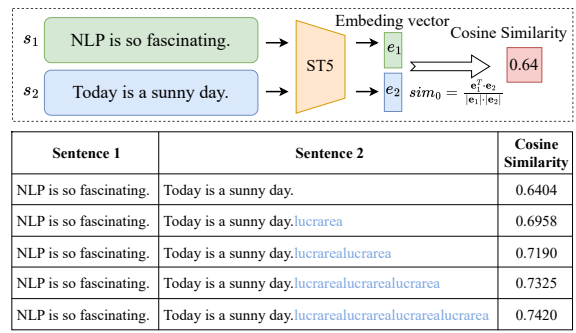

Figure 1: An example illustrating how a sticky token affects sentence cosine similarity in text embedding models. Inserting the token multiple times makes two sentences appear more similar than they actually are.

In recent years, Transformer-based embedding models have become increasingly prominent due to their high performance, including Sentence-BERT (Reimers and Gurevych, 2019a), Sentence-T5 (Ni et al., 2021a), and others. Recently, the community started to fine-tuning decoder-only LLMs for embedding (BehnamGhader et al., 2024; Muennighoff et al., 2024; Yavuz et al., 2024). Crucially, all these models depend on tokenization to convert text into subword units — tokens.

Despite ongoing efforts to refine tokenization algorithms (Sennrich et al., 2016; Kudo and Richardson, 2018; Kudo, 2018; Schmidt et al., 2024), anomalous token behaviors still emerge. For example, "glitch tokens" (BehnamGhader et al., 2024; Muennighoff et al., 2024) can exhibit unintended effects on language model outputs. More recently, Kaggle (2024) reported another surprising behavior in *text embedding model*: inserting certain tokens can make two sentences appear more similar than they actually are. As illustrated in Figure 1, repeatedly appending the token "lucrarea" to an unrelated sentence yields a noticeable increase in its similarity to a reference sentence when using Sentence-T5 (ST5) (Ni et al., 2021a). This suggests the existence of a novel class of anomalous tokens that not only alters embedding distributions but also can degrade downstream performance in real-world tasks. However, no systematic study has yet investigated how these tokens operate, how to detect them, and how they affect embedding-based applications.

In this paper, we conduct an in-depth exploration of these unusual "sticky tokens". Through preliminary experiments, we find that while such tokens sometimes raise similarity between sentences, their primary tendency is to "pull" sentence pairs toward a particular similarity value—often the mean similarity in the model's token-embedding space. Consequently, they reduce variance in similarity without regard to the underlying semantics of the texts.

To rigorously investigate this phenomenon, we formally define "sticky tokens" and propose an efficient detection approach, **S**ticky **T**oken **D**etector (**STD**), based on filtering both sentence pairs and candidate tokens. We apply STD to 40 models spanning 14 prominent model families and uncover a total of 868 sticky tokens. Our results reveal that sticky tokens frequently stem from *special* or *unused* tokens, as well as subword fragments in multiple languages; their prevalence does not strictly correlate with model size or vocabulary size. Furthermore, we show that inserting these tokens causes notable performance degradation in downstream tasks: for instance, retrieval accuracy on NFCorpus can fall by over 50% for certain models. A layer-wise attention analysis suggests that sticky tokens disrupt normal attention patterns, overshadowing other parts of the input sequence.

Our findings highlight a largely overlooked tokenization issue in text embedding models. We hope this work will spark future research on designing robust tokenizers and model architectures that mitigate the effects of sticky tokens, ultimately leading to more reliable embedding-based NLP systems.

## 2 Related work

**Tokenization** Tokenization is a crucial process in modern NLP systems, yet it can also introduce problematic behaviors (Phan et al., 2024; Wang et al., 2024a; Singh and Strouse, 2024; Mielke et al., 2021). Popular subword tokenization methods, including Byte-Pair Encoding (BPE) (Sennrich et al., 2016), WordPiece (Kudo and Richardson, 2018), and Unigram (Kudo, 2018), have been widely adopted in large-scale text processing pipelines. Despite their advantages in handling vocabulary size and rare words, these methods can still yield undesirable outcomes, such as splitting meaningful terms into unintuitive fragments or creating tokens that rarely occur in the training data (Karpathy, 2024; Chai et al., 2024).

**Anomalous Token** Recent research on LLMs has highlighted a variety of unexpected token-level anomalies. For instance, Land and Bartolo (2024a) identify "under-trained" tokens in LLMs, while Li et al. (2024), Zhang et al. (2024), and Wu et al. (2024) investigate so-called "glitch tokens" that exhibit abnormal behaviors due to incomplete or skewed pre-training coverage. These studies explore detection methods and propose strategies to mitigate the harmful effects of such tokens on language model outputs. However, their primary focus lies in LLMs, leaving the anomaly space of *text embedding models* largely unexplored.

## 3 Problem Formulation

In this section, we first explore how certain anomalous tokens differ from normal tokens by observing their influence on sentence similarity. Then, based on our findings, we formally define these tokens.

### 3.1 Anomalous Behavior

Certain tokens have been identified in previous work (Kaggle, 2024) as behaving unusually. For example, </s> and lucrarea in the ST5-base model were reported to increase pairwise sentence similarity in some cases as shown in Figure 1. However, beyond these observations, there has been no detailed or systematic analysis of such anomalous tokens.

Figure 2 shows a typical example of this behavior in the ST5-base model. We randomly sampled 1,000 sentences from Wikipedia and computed pairwise cosine similarity. We then selected sample pairs at intervals of 0.02 (from the sorted similarity value list) and added either a normal token (e.g., and; Figure 2a) or an anomalous token (e.g., lucrarea; Figure 2b) to one sentence in each pair, repeating the token multiple times. Additionally, Figure 2c shows the density distribution of sentence similarities in ST5-base's embedding space. Following previous work (Gao et al., 2019; Fuster Baggetto and Fresno, 2022), we use token embeddings as a surrogate for text embeddings since both share the same embedding space.

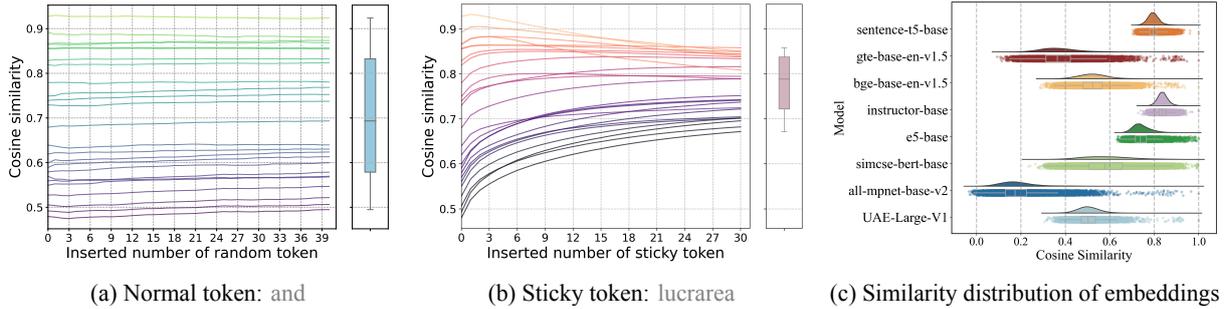

(a) Normal token: and    (b) Sticky token: lucrarea    (c) Similarity distribution of embeddings

Figure 2: Sentence similarity trend curves and similarity distributions for various tokens and text embedding models. (a) and (b) compare the impact of adding multiple occurrences of a normal token (and) vs. a sticky token (lucrarea) to one sentence in each randomly selected sentence pair using the ST5-base model. We sample sentence pairs from Wikipedia, compute their pairwise similarity, then plot how the similarity changes as we add more tokens. The line plots show the relationship between the number of added tokens and sentence cosine similarity, while the boxplots show the quartiles of the final similarity values. (c) displays the distribution of cosine similarity between token/word-embeddings for different models. We use token embeddings as a surrogate for text embeddings since both share the same embedding space. See Appendix A for more examples of other tokens and results on additional models.

Our results reveal that repeatedly adding the anomalous token lucrarea consistently "pulls" the pairwise similarity to a value near the *median* of the distribution, which also aligns with the *mean* pairwise similarity among token embeddings for ST5-base (Figure 2c).[1]

On the other hand, adding a normal token like and has a much smaller impact on sentence similarity (Figure 2a). Interestingly, although anomalous tokens can sometimes increase sentence similarity (as noted in previous observations), this does not always happen. Their influence does not have to be strictly monotonic, and not all sentence pairs are affected in the same way.

See Appendix C for our conjecture to explain this anomalous behavior.

### 3.2 Formalization

Let $E : \mathbb{S} \to \mathbb{R}^d$ be a text embedding model mapping a sentence $s \in \mathbb{S}$ to a $d$-dimensional vector $E(s)$. We can write $\mathbb{S}$ as $\mathbb{V}^m$, where $\mathbb{V}$ is the set of all tokens in the vocabulary. We measure the similarity between embeddings using cosine similarity, defined as: $Sim(\boldsymbol{u}, \boldsymbol{v}) = \boldsymbol{u}^\top \boldsymbol{v}/\|\boldsymbol{u}\|\|\boldsymbol{v}\|$ (i.e. the dot product between $\ell_2$ normalized $\boldsymbol{u}$ and $\boldsymbol{v}$). Let $Sim(s_1, s_2)$ denote the similarity between $E(s_1)$ and $E(s_2)$.[2] Higher values of $Sim(\cdot, \cdot)$ indicate greater similarity.

Anomalous tokens are first noticed when inserted into existing sentences (Kaggle, 2024). Inserting a token $t$ into a sentence $s$ can happen in different ways, including (1) repeatedly adding $t$ at the beginning (prefix), (2) repeatedly adding $t$ at the end (suffix), or (3) adding $t$ at random positions[3]. We denote these operations with $\mathbb{I} = \{\mathcal{I}_{\text{pre}}, \mathcal{I}_{\text{suf}}, \mathcal{I}_{\text{ran}}\}$. Each $\mathcal{I} \in \mathbb{I}$ takes as input $(s, t, n)$ and produces a new sentence containing $n$ insertions of $t$ at positions determined by the specific insertion operation.

As shown in Figure 2b, anomalous tokens tend to pull sentence similarity toward the mean of the model's token-similarity distribution if they are inserted repeatedly. In other words, they reduce the distance between the pairwise similarity of two arbitrary sentences and this mean value, or they decrease the variance of that similarity distribution. We name these *sticky tokens* and formally define them as follows:

**Definition 1.** *Given a text embedding model $E$ and $u$, the mean pairwise similarity of its token embeddings, a token $t$ is called a* sticky token *if, for all $s_1, s_2 \in \mathbb{S}$ and for all $\mathcal{I} \in \mathbb{I}$, we have:*

$$\left|Sim(s_1, \mathcal{I}(s_2, t, n)) - u\right| \leq \epsilon.$$

Here, $n$ and $\epsilon$ are parameters chosen based on how much change in sentence similarity is considered significant by the model's users. Different values of $\epsilon$ or $n$ will identify different sets of sticky tokens. In practice, $\mathbb{S}$ should be large and diverse, covering varied syntactic structures, semantic meanings, and domain contexts to ensure evaluation robustness.

---
[1] In Figure 2b, the *median* of the sentence similarity curve for ST5-base is about 0.8. This matches the *mean* pairwise similarity of the model's token embeddings (also around 0.8) shown in Figure 2c.

[2] For clarity, "sentence similarity" and "token similarity" both refer to comparisons in the embedding space in this paper.

[3] See Appendix B for other insertion operations.

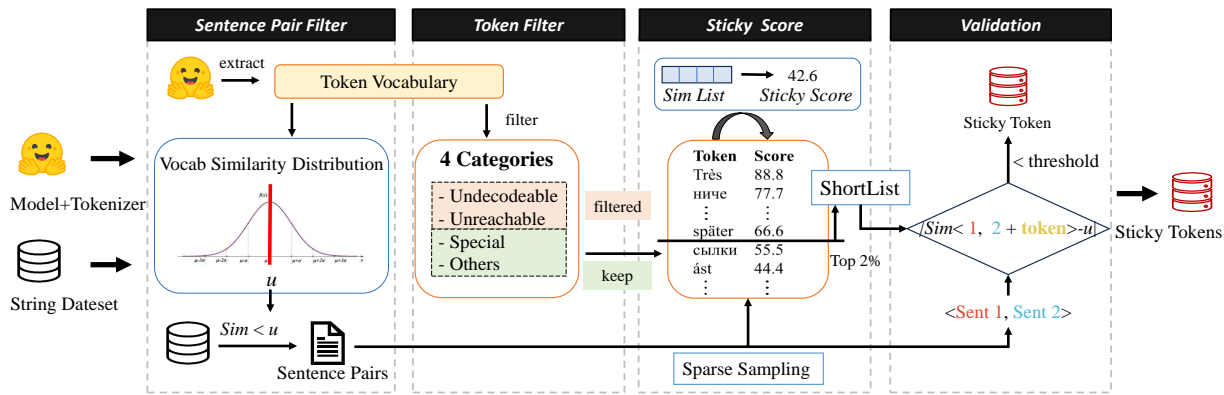

Figure 3: The framework of STD to detect sticky tokens.

## 4 Methodology

Based on the concept of sticky tokens in Definition 1, we propose STD to detect these tokens in a given text embedding model. As shown in Figure 3, STD takes two inputs: the target text embedding model and a set of sentences. It outputs a list of sticky tokens from the model's vocabulary.

A direct application of Definition 1 would require checking pairwise-similarity changes for every possible sentence pair and every token, which can be very costly. However, the examples in Figure 2 suggest that focusing on only part of the sentence pairs is enough to distinguish sticky tokens from normal ones. For instance, sticky tokens usually pull sentence similarity towards the overall *mean* of the model's token-similarity distribution (especially for sentence pairs with initial similarity below that *mean*). Building on this insight, we adopt a more efficient detection strategy with four main steps:

1. *Sentence Pair Filtering*: Filter out sentence pairs whose initial similarity is already above the mean of the distribution.

2. *Token Filtering*: Remove tokens that are undecodable, unreachable, or otherwise invalid.

3. *Shortlisting via Sticky Scoring*: Compute a "sticky score" for each candidate token to create a shortlist.

4. *Validation*: Verify that the shortlisted tokens truly satisfy the formal definition of a sticky token (Definition 1).

### 4.1 Sentence Pair Filtering

Figure 2 shows that sticky tokens have a clear impact on sentences whose initial similarity is below the mean similarity ($u$) of the token embedding space.[4] To reduce the search space, we only keep those pairs $(s_1, s_2)$ in the set $\mathbb{S}$ s.t.

$$Sim(s_1, s_2) < u.$$

We call this filtered set $\mathbb{P}_f$. By focusing on sentence pairs with relatively lower similarity, we can check whether a token consistently increases their similarity, bringing it closer to $u$.

### 4.2 Token Filtering

Following previous work (Land and Bartolo, 2024b), we also remove certain tokens that the model cannot decode or handle properly. In particular, we categorize tokens in models' vocabulary $\mathcal{V}$ into the following categories:

- **Undecodable tokens**: These contain invalid characters or cannot be decoded into readable text.

- **Unreachable tokens**: These cannot be reproduced by decoding and re-encoding (the token ID changes and is not mapped back to the original ID).

- **Special tokens**: These are tokens used by the model for special purposes (e.g., [CLS], [SEP], or </s>).

- Others: Tokens not in any of the other categories, which constitute the vast majority.

---
[4]The way we compute $u$ is discussed in Appendix D.1.

We filter out *undecodable* and *unreachable* tokens from our sticky token detection pipeline.[5] We denote the remaining valid token set as $\mathcal{V}^*$, which we use in the following steps.

### 4.3 Shortlisting Tokens with Sticky Scores

After filtering the sentence pairs and the vocabulary, we need to identify which tokens in $\mathcal{V}^*$ behave like sticky tokens. A naive way to do this would be to test each token on every pair in $\mathbb{P}_f$, but that can still be expensive. Instead, we work with a smaller, randomly sampled subset of $\mathbb{P}_f$ to compute a "sticky score" that helps us shortlist the most likely sticky tokens.

**Measuring Influence.** Suppose we have $k$ sampled sentence pairs,

$$p_j \in \mathbb{P}_f, \quad p_j = (s_1^j, s_2^j),$$

and let $\mathcal{I}$ be an insertion operation (e.g., prefix, suffix, or random insertion). For each token $t$, we insert it multiple times into one sentence of the pair $(s_1^j, s_2^j)$. We then calculate

$$\Delta_{(t,\mathcal{I})}^j = Sim(s_1^j, \mathcal{I}(s_2^j, t, n)) - Sim(s_1^j, s_2^j).$$

This value $\Delta_{(t,\mathcal{I})}^j$ represents how much the similarity changes when token $t$ is inserted.

**Sticky Score.** We summarize these changes across all sampled sentence pairs in two ways:

- $M_{(t,\mathcal{I})}^+$: the total amount of positive changes in similarity.
- $M_{(t,\mathcal{I})}^-$: the total amount of negative changes in similarity.

We also track the frequencies $F^+$ and $F^-$, which are the percentages of pairs that show positive and negative changes, respectively. Finally, we include $Sim(s_1, t)$ to account for how semantically close $t$ is to the sentence (which might inflate similarity artificially).

Putting these together, we define a sticky score:

$$\mathcal{SS}_{\mathcal{I}}(t) = \frac{M_{(t,\mathcal{I})}^+ + \alpha\, F_{(t,\mathcal{I})}^+}{M_{(t,\mathcal{I})}^- + \beta\, F_{(t,\mathcal{I})}^- + Sim(s_1, t) + \gamma},$$

where $\alpha$ and $\beta$ are small positive constants to balance magnitude and frequency, and $\gamma$ is a small constant included for numerical stability. Then

---

[5]See Appendix D.2 for more details on these categories.

we aggregate across all insertion operations and sampled pairs to get a final score:

$$\mathcal{SS}(t) = \sum_{\mathcal{I} \in \mathbb{I}} \sum_{p \in \mathbb{P}_f} \mathcal{SS}_{\mathcal{I}}(t).$$

Tokens that rank in the top 2% of $\mathcal{SS}(t)$ form our shortlist of potential sticky tokens.

### 4.4 Validation of Shortlisted Tokens

Finally, we check each shortlisted token to confirm it meets the formal definition of a sticky token (Definition 1). Here, we use all sentence pairs in $\mathbb{P}_f$ rather than just a small subset. As shown in Algorithm 1, each candidate token is inserted into many pairs in multiple ways (prefix, suffix, or random). We then measure whether the distance to $u$ remains below a threshold $\epsilon$, reflecting that the token truly "pulls" similarity to that mean.

Since different embedding models have different value ranges and distributions, we propose an adaptive threshold in Algorithm 2 to pick $\epsilon$. This helps adjust to model-specific characteristics and ensures that detected tokens really exhibit the distinctive behavior of sticky tokens.

## 5 Evaluation

In this section, we apply STD (Section 4) to find sticky tokens in well-known text embedding models. We also examine how the presence of these tokens affects downstream tasks and investigate potential reasons for their anomalous behavior.

### 5.1 Evaluation Setup

**Dataset.** We use the Semantic Textual Similarity (STS) datasets as our collection $\mathbb{S}$, since they naturally include sentence pairs. Specifically, we take STS datasets from the Massive Text Embedding Benchmark (MTEB)[6] (Muennighoff et al., 2023), including STS12, STS13, STS14, STS15, STS16, STS17, STS22, STSBenchmark, BIOSSES, and SICK-R (Agirre et al., 2012, 2013, 2014, 2015, 2016).

**Target Text Embedding Model.** We evaluate a diverse range of 12 text embedding model families published between 2019 and 2025, including Sentence-BERT (Reimers and Gurevych, 2019b), SimCSE (Gao et al., 2022), Sentence-T5 (Ni et al., 2021a), GTR (Ni et al., 2021b), Instructor (Su et al., 2023), E5 (Wang et al., 2024b,c), BGE (Xiao et al., 2024), AnglE (Li and Li, 2024), Nomic (Nussbaum et al., 2025), GTE (Li et al., 2023), jina (Sturua et al., 2024), KaLM (Hu et al., 2025), GritLM (Muennighoff et al., 2024), and SFR (Yavuz et al., 2024). A detailed overview of each model is given in Appendix E.

---

[6]https://huggingface.co/mteb

| Model | Model Size | Vocab Size | Filter Passed | Candidate | Validated | Examples |
|---|---|---|---|---|---|---|
| all-MiniLM-L6-v2 | 23M | 30522 | 23699 | 474 | 21 | (, h₂o, [CLS], ₂,gambia |
| all-mpnet-base-v2 | 109M | 30527 | 23700 | 474 | 24 | 00 ,た, 「, т, ← |
| sup-simcse-bert-base-uncased | 109M | 30522 | 23699 | 474 | 22 | 203, ?, [SEP], 口, り |
| sup-simcse-bert-large-uncased | 335M | 30522 | 23699 | 474 | 11 | ', ;, contestants, accidental, ɔ, ] |
| sup-simcse-roberta-base | 125M | 50265 | 49894 | 998 | 27 | Ġthere, </s>, âĠĵâĠĵ, ĠâĠĶ, ĠÂŃ, .âĠĶ, ÂŃ, Ġï¿½, âĠĺ |
| sup-simcse-roberta-large | 355M | 50265 | 49894 | 998 | 15 | ĠâĠĭ, Ġ?, .-, Ġschematic, )]. |
| sentence-t5-base | 110M | 32100 | 32097 | 642 | 21 | </s>, lucrarea,<extra_id_18>, ▁grains, ▁photographed |
| sentence-t5-large | 336M | 32100 | 32097 | 642 | 30 | </s>, ▁»., <extra_id_27>, ▁Comment, ▁Ribbon |
| sentence-t5-xl | 1242M | 32100 | 32097 | 642 | 34 | </s>, <extra_id_0>, <extra_id_27>, ▁velvet, ▁context |
| sentence-t5-xxl | 4866M | 32100 | 32097 | 642 | 22 | </s>, ▁consacré, <extra_id_27>, ▁hashtag, ▁hello |
| gtr-t5-base | 110M | 32100 | 32097 | 642 | 16 | </s>, lucrarea, ▁Someone, <extra_id_26>, ▁happened |
| gtr-t5-large | 336M | 32100 | 32097 | 642 | 14 | ▁»., </s>, <extra_id_27>, <extra_id_25>, ▁supposed |
| gtr-t5-xl | 1242M | 32100 | 32097 | 642 | 15 | </s>, <extra_id_0>, <extra_id_9>, <extra_id_27>, ▁badly |
| gtr-t5-xxl | 4866M | 32100 | 32097 | 642 | 7 | </s>, ▁consacré, ▁shortly, Pourtant, ▁indeed |
| instructor-base | 110M | 32100 | 32097 | 642 | 12 | </s>, lucrarea, <extra_id_26>, ▁somewhere, <extra_id_19> |
| instructor-large | 336M | 32100 | 32097 | 642 | 32 | </s>, ▁»., <extra_id_27>, ▁waiting, ▁exhausted |
| instructor-xl | 1242M | 32100 | 32097 | 642 | 8 | </s>, <extra_id_0>, <extra_id_9>, <extra_id_27>, ▁newly |
| e5-small | 33M | 30522 | 23699 | 474 | 17 | [SEP], exhibiting, occurring, pretended, behaved |
| e5-base | 109M | 30522 | 23699 | 474 | 21 | generating, absorbing, heating, carpet, human |
| e5-large | 335M | 30522 | 23699 | 474 | 21 | ⇄, ǂ , [SEP], ∅, 王 |
| e5-mistral-7b-instruct | 7111M | 32000 | 31747 | 635 | 31 | ▁sont, ▁peut, ▁много, ▁жду, ▁испо |
| bge-small-en-v1.5 | 33M | 30522 | 23699 | 474 | 18 | [, m³, ð, [PAD], [SEP] |
| bge-base-en-v1.5 | 109M | 30522 | 23699 | 474 | 20 | neighbouring, ? , witnessed, granting, 。 |
| bge-large-en-v1.5 | 335M | 30522 | 23699 | 474 | 15 | actively, intended, intercepted, intentional, uploaded |
| UAE-Large-V1 | 335M | 30522 | 23699 | 474 | 14 | [SEP], ɔ, n, occurring, having |
| nomic-embed-text-v1 | 137M | 30522 | 23699 | 474 | 12 | [CLS], [MASK], ¦ , polling, 勝 |
| nomic-embed-text-v1.5 | 137M | 30522 | 23699 | 474 | 9 | [CLS], [MASK], [SEP], cerambycidae, ∼ |
| gte-small | 33M | 30522 | 23699 | 474 | 15 | [SEP], [CLS], treacherous, 2nd, peacefully |
| gte-base | 109M | 30522 | 23699 | 474 | 18 | [SEP], [MASK], hotspur, [CLS], aroused |
| gte-large | 335M | 30522 | 23699 | 474 | 18 | 1 ,⸝st, 30th, mcgrath, rendering |
| gte-base-en-v1.5 | 137M | 30522 | 23699 | 474 | 20 | [CLS],[PAD], ∞, ₃, ■, ⊕, ⇄, ⊼, ℓ, ∩, 王, ⸜ |
| gte-large-en-v1.5 | 434M | 30522 | 23699 | 474 | 17 | ǂ , multiplied, ∴, ∧, z |
| gte-Qwen2-1.5B-instruct | 1543M | 151643 | 147848 | 2326 | 5 | Ġthru, Ġgifted, Ġupfront, Ġportraying, Ġawkward |
| gte-Qwen2-7B-instruct | 7069M | 151643 | 147848 | 2957 | 103 | Ġanon, Ġcommenting, Ġsolver, ĠChecking, ĠSteering |
| jina-embeddings-v3 | 572M | 250002 | 249976 | 5000 | 40 | </s>, <s>, ╀, Ґ, 嚓, 今次, 出, イ, ム |
| KaLM-instruct-v1 | 494M | 151643 | 147848 | 2957 | 27 | astically, ×©×ķ×ĺ×¨, piÄĻ, alty, czyÄĭ, ×ĺ×ij×ĺ |
| KaLM-instruct-v1.5 | 494M | 151643 | 147848 | 2957 | 31 | versible, ×Ļ,×Ĺ×ª, (bounds, afety, ×¢×ĺ×Ķ, ÃjP |
| GritLM-7B | 7111M | 32000 | 31747 | 635 | 17 | ▁adventures, ▁promoting, ▁nine, ▁folks, ▁village |
| SFR-Embedding-2_R | 7111M | 32000 | 31716 | 444 | 2 | zeichnet, ▁scales |
| SFR-Embedding-Mistral | 7111M | 32000 | 31716 | 635 | 46 | ▁которы, ▁годи, ▁Jahrhund, ▁который, ▁которых |

Table 1: Statistics and validated sticky tokens of target models. The column **Validated** represents the number of validated sticky tokens. **Examples** are manually chosen based on readability, similarity across the models, and also representativeness. Note that some leading characters (e.g., ▁ or Ġ) are utilized by tokenizers to indicate spaces or word boundaries.

**Hyperparameter.** From Definition 1 and Section 4.3, we must choose (i) $n$, the number of times each token is inserted, (ii) $k$, the number of sentence-pair samples to obtain sticky score, and (iii) $\epsilon$, the threshold for verifying stickiness. Through ablation studies[7], we pick $n = 8$ and $k = 5$. Specific threshold values $\epsilon$ for each model are also provided in Table 7 in Appendix E.

---
[7]See Appendix E for detection experiment details.

### 5.2 Experimental Results

Table 1 lists each model's size, along with the number of detected sticky tokens and a few token examples. We first discuss general trends, followed by observations unique to specific model families.

#### 5.2.1 General Observations

We discover a total of 868 sticky tokens across 40 model checkpoints. The number of verified sticky tokens depends on both the model family and the size of the tokenizer's vocabulary. Overall, the percentage of sticky tokens (among all shortlisted candidates) ranges from 0.4% to 5.3%, corresponding to 0.006% to 1% of the total vocabulary. This suggests that STD and shortlisting steps are efficient.

We also find that the forms of these tokens vary significantly among different model families:

- *Models from the same family* often share sticky tokens.
- There is no direct or consistent correlation between model size/vocabulary size and the number of sticky tokens.
- *Unused* or *special* tokens frequently appear in the sticky token set.

Below are some more specific examples.

**Special and Control Tokens.** Many models include special tokens for certain functionalities, such as marking start/end of sequences or separating segments. We observe that:

- About 7% (64 tokens) of the 868 sticky tokens belong to this category, including </s>, [CLS], [SEP], [MASK], [PAD].
- Some *unused* tokens[8] (e.g., <extra_id_18>, <extra_id_27>) also appear as sticky tokens.
- Certain tokens like </s> and <extra_id_27> show up many times (12 and 8, respectively) across multiple T5-based checkpoints.

These observations hint that special tokens might unintentionally confuse the model's embedding space, although the reasons remain to be explored in future work.

**Multilingual and Non-ASCII Fragments.** About 22% (191 tokens) of detected sticky tokens contain characters beyond the standard English alphabet. Examples include:

- Cyrillic fragments (т, х, ра, ци),
- CJK tokens (う, 治, 水, 口),
- Arabic subwords (ـ, ﺟ),
- combining diacritics (՚, ׃, °),
- mathematical symbols (³, ∩, ∞).

In many cases, these tokens appear as single characters or subword segments detached from their usual context, likely due to multilingual training data and Byte Pair Encoding (BPE). For instance, _ч (a Cyrillic prefix) and _släktet (Swedish for "the genus") may lose important contexts. This suggests *sticky tokens may emerge from limited non-English coverage during pre-training*.

---

[8]Unused tokens are reserved tokens in pretrained models' vocabularies that weren't utilized during pretraining (Land and Bartolo, 2024a).

### 5.2.2 Model-Specific Observations

This section presents model-specific observations on sticky tokens. Our analysis reveals variations in the prevalence and characteristics of sticky tokens across various models, underscoring the influence of tokenizer design and model scale.

**T5-Based Models.** The T5 family (sentence-t5, gtr-t5, instructor) exhibits consistent patterns associated with its SentencePiece tokenizer (Kudo and Richardson, 2018) (vocab_size=32,100). All variants include the end-of-sequence token </s> as a sticky token. Larger T5 models show a non-linear correlation between the number of parameters and the frequency of sticky tokens (The spearman's correlation analysis is: $\rho = 0.127$, p-value = 0.706). For instance, sentence-t5-xl (1.2B) contains 34 sticky tokens, the highest among T5 variants, while sentence-t5-xxl (4.8B) reduces this to 22. Some unused tokens (e.g., <extra_id_27> in 8 out of 11 T5-based models) and non-English fragments (lucrarea, _consacré), appear frequently in sticky token lists. These may be residuals from the model's pre-training phase. Notably, instructor-xl (1.2B) shows the lowest sticky token count (8 tokens), suggesting improved token robustness after post-training adjustments.

**BERT/RoBERTa Derivatives.** Models using BERT-style tokenizers (Devlin et al., 2018; Liu et al., 2019) (vocab_size ≈ 30k–50k) exhibit an inverse correlation between sticky token counts and model parameter size. For example, sup-simcse-bert-large-uncased (335M) contains only 11 sticky tokens (e.g.,՚,;,ɔ), while all-mpnet-base-v2 (109M) has 24 sticky tokens. RoBERTa models (Liu et al., 2019) display distinct characteristic: sup-simcse-roberta-base (125M) includes 27 sticky tokens, primarily consisting of malformed subwords (e.g., âĢjâĢĵ, ĠâĢĶ), while its 355M-parameter counterpart includes only 15 sticky tokens, retaining punctuation-related tokens such as Ġ?) and .).

**LLM-based Models.** Other LLM-based Models with 7B parameters show notable variations on the number of sticky tokens. For example, GritLM-7B exhibits common sticky token counts (17, e.g., _adventures, _young), while gte-Qwen2-7B-instruct stands out with 103 sticky tokens, the highest count observed, including frequent verb participles (Ġcommenting, Ġfixing) and technical terms (Ġsyncing, Ġtaxable). In contrast, SFR-

Embedding-Mistral (7B) encounters significant problems in processing non-English tokens. For example, 46 sticky tokens of it are composed of Cyrillic subwords (_которы). These observations suggest that *there is no consistent pattern between the presence of sticky tokens and model scale or vocabulary size.*

**Multilingual and Domain-tuned Models.** Multilingual models alwasys reveal cross-script vulnerabilities. For example, E5-mistral-7B-instruct contains 31 sticky tokens across 7 scripts (e.g., Cyrillic _ст, Hebrew γ). Smaller models, such as UAE-Large-V1 (335M), have problems on script-specific partial tokens (e.g., i, ʊ, א). Domain-tuned models show task-specific issues. For example, medical terms like Cerambycidae appear as sticky tokens of nomic-embed-text-v1.5 while numerical ordinal tokens (e.g., 3a, 55th) frequently appear in the sticky token list of GTE-family models. These findings indicate that *multilingual capabilities and domain-specific fine-tuning may lead to the emergence of sticky tokens.*

### 5.3 Impact on Downstream Tasks

This section we aim to investigate the impact of sticky tokens on downstream tasks.

**Method.** We use a curated 15-task subset from MTEB benchmark (Muennighoff et al., 2023) as the datasets. For each model, we insert previously verified sticky tokens (Section 5.2) or randomly chosen normal tokens into sentences or paragraphs within the datasets. We use the metrics from MTEB (Muennighoff et al., 2023) for comparison. See Appendix F for datasets and method details.

**Results** Table 2 shows the partial results[9] of our evaluation on clustering and retrieval tasks. Compared with normal tokens, sticky tokens demonstrate significantly higher destructiveness (Paired t-test results: t = 2.23, p = 0.017; one-tailed; mean difference (normal-sticky) = 2.23; Cohen's d = 0.41). This confirms their greater destructiveness at p < 0.05. For instance, for the ST5-base model, inserting normal tokens shows minimal degradation (SciFact: 45.76→44.58, Δ-2.6%; NFCorpus:28.64→28.48, Δ-0.56%), while inserting sticky tokens cause a significant degradation (SciFact: 45.76→26.76, Δ-41.5%; NFCorpus: 28.64→13.65, Δ-52.3%). Furthermore, lightweight models suffer catastrophic degradation from sticky tokens (sentence-t5-base on Biorxiv clustering: 23.11→15.02, Δ-35.0%), while larger models like UAE-Large-V1 maintain robustness (SciFact retrieval: 73.91→72.63, Δ-1.7%). Our experiments reveal that *sticky tokens significantly degrade performance across downstream tasks.*

---
[9]See Table 9 in Appendix F for the full downstream results.

| Categories → | Clustering | | | Retrieval | | |
|---|---|---|---|---|---|---|
| Datasets → | Biorxiv Clustering | Medrxiv Clustering | TwentyNewsgroups Clustering | SciFact | ArguAna | NFCorpus |
| sentence-t5-base | 23.11 | 26.03 | 49.27 | 45.76 | 44.84 | 28.64 |
| w/ normal token | 20.04 | 25.06 | 37.17 | 44.58 | 45.41 | 28.48 |
| w/ sticky token | **15.02** | **20.41** | **35.38** | **26.76** | **42.14** | **13.65** |
| instructor-base | 26.40 | 28.38 | 52.77 | 57.88 | 51.18 | 30.76 |
| w/ normal token | 18.05 | 23.13 | 50.64 | 57.70 | 47.45 | 29.77 |
| w/ sticky token | 26.05 | 26.55 | **50.55** | **43.47** | **47.03** | **23.11** |
| e5-base | 29.92 | 27.67 | 43.75 | 71.88 | 53.03 | 37.09 |
| w/ normal token | 28.94 | 26.51 | 22.15 | 71.36 | 51.13 | 37.15 |
| w/ sticky token | **27.02** | **24.92** | **20.00** | **70.95** | **49.14** | **37.01** |
| simcse-bert-base | 25.70 | 25.85 | 31.67 | 33.89 | 39.56 | 13.49 |
| w/ normal token | 25.11 | 25.19 | 28.40 | 33.66 | 36.79 | 13.45 |
| w/ sticky token | **24.80** | **25.17** | 29.22 | **29.89** | 38.38 | **8.84** |
| UAE-Large-V1 | 37.24 | 31.18 | 51.72 | 73.91 | 66.15 | 37.61 |
| w/ normal token | 35.79 | 30.96 | 40.48 | 74.51 | 63.67 | 37.70 |
| w/ sticky token | 35.98 | **30.94** | 47.20 | **72.63** | **63.48** | 37.79 |

Table 2: Results on Downstream Tasks. We present the performance of four models, comparing their baseline results with sticky tokens and normal tokens.

### 5.4 Explainability of Causes

We conduct a preliminary analysis to explore the underlying causes of the sticky token phenomenon. We compare the observed attention patterns and analyze layer-wise divergence between sticky tokens and normal tokens. Experiments are conducted on 1k Wikipedia sentences appended with either sticky tokens (e.g., </s>) or normal tokens (random selected). Here, we present the results obtained with the ST5-base model.

**Attention Pattern Disparity.** For each sequence and attention head, the attention weights at the position of the added token are extracted from the corresponding column vector of the attention score matrix[10]. This reflects how the token is attended to by the others in the sequence. As illustrated in Figure 4 left, when sticky tokens are appended to sentences, their attention weights in intermediate layers concentrate disproportionately in high-value ranges (e.g., weights>0.4), whereas normal tokens follow a smoother, more Gaussian/Normal distribution. This suggests that *sticky tokens dominate the model's attention and disrupt the balanced contextual representation of input texts*.

**Layer-Wise Amplification of Anomalies.** The Wasserstein distance (Vaserstein, 1969) between the attention patterns of sticky and normal tokens (Figure 4 right) further elucidates how anomalies propagate across layers. In early layers (1–6), the divergence remains moderate, indicating that shallow processing retains some robustness. However, from mid to late layers (6–12), the distance increases, peaking at the final layers. This reflects a compounding effect: minor irregularities in early layers are progressively amplified as deeper layers integrate higher-order semantic features.

---
[10]See Appendix G for details to obtain attention pattern.

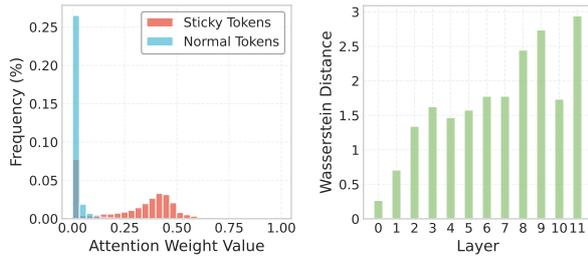

Figure 4: The example distribution of attention patterns (left) and Wasserstein distance of the probability distributions between sticky tokens and normal tokens in different intermediate layers of ST5-base model (right). Sticky tokens (red) exhibit higher frequency in high-attention regions (>0.4) compared to normal tokens (blue).

For text embedding models, the amplification disrupts the hierarchical abstraction of semantics. *The anomalous intermediate results caused by sticky tokens are not uniformly distributed across all layers of the model but are concentrated and amplified in specific key layers.*

## 6 Conclusion

In summary, STD successfully detects 868 sticky tokens in 40 text embedding models and demonstrates that these tokens can significantly degrade downstream performance on tasks such as clustering and retrieval. Through comprehensive experiments, we show that sticky tokens often stem from special or unused tokens and subword fragments from multiple languages, suggesting that tokenizer design and pre-training coverage both play important roles. We further provide evidence of how these tokens cause anomalies in the attention layers, amplifying small irregularities into major distortions of final sentence representations. Our findings encourage future work on designing more robust tokenization schemes and model architectures to mitigate the effect of sticky tokens.

## Limitations

Although our definition of sticky tokens is as detailed as possible, and our pipelines for detecting sticky tokens on different models are also effective, they still have some significant limitations.

Most notably, we assume that sticky tokens uniformly "pull" similarity toward the token embedding mean. However, models with non-Gaussian token similarity distributions(Model with isotropic embedding space [11]) (Li et al., 2020; Su et al., 2021) or task-specific embeddings might require tailored detection criteria. It remains unclear whether these models exhibit abnormal features akin to sticky token properties. Future research on model interpretability could refine our deeper understanding of model embedding space and sticky token phenomenon, and lead to more effective detection methods.

Secondly, while we identify the anomalous phenomenon and its downstream impacts, we do not propose concrete solutions to mitigate sticky tokens (e.g., tokenizer retraining, embedding space regularization). Our experiments involve inserting tokens at fixed positions (prefix, suffix, or random) with a predefined repetition count. While we also examined why alternative insertion methods, such as deletion or replacement, were not incorporated（Appendix B）, our analysis did not extend to more complex adversarial scenarios. These scenarios could include advanced strategies like interleaving tokens or context-aware placement, which were not evaluated in our study.

Finally, the scope of detection of our work is limited to focusing on open source text embedding models, which often use byte-pair encoding based tokenization. However, the detection results may differ for certain closed-source models, such as OpenAI's text-embedding series or Google's gemini-embedding series, as well as models using Unigram-based tokenization. Additionally, obtaining the vocabulary for these closed-source models presents a significant challenge.

## Acknowledgements

This work is supported by the State Key Laboratory of Industrial Control Technology, China (Grant No.ICT2024C01), and the Fundamental Research Funds for the Central Universities, China (Grant No.2025ZFJH02). Additionally, the authors wish to thank the anonymous reviewers for their helpful comments.

---

[11]See Appendix C for discussion about anisotropy vs. isotropy of embedding model. Note that we evaluated 40 widely-used embedding models, and all of them exhibited anisotropic behavior.

## A Symptom Across Models

The examples shown in Figure 2 in Section 3.1 were drawn from extensive empirical testing. Due to space constraints, only a few representative cases were included in the main text. Additional examples are available in our repository.

Similar to the issue of anomalous tokens observed in the ST5-base model with lucrarea, we provide further random examples of these anomalous tokens across various models to illustrate the prevalence of this phenomenon across different models. The experimental setup remains consistent with Section 3.1: we randomly sampled 1,000 sentences from Wikipedia and computed pairwise cosine similarity. We then selected sample pairs at intervals of 0.02 (from the sorted similarity list) and added anomalous token to one sentence in each pair, repeating the token multiple times. We found that repeatedly adding the anomalous token consistently "pulls" the pairwise similarity to a value near the *median* of the distribution, which also aligns with the *mean* pairwise similarity among token embeddings for corresponding model(Figure 10).

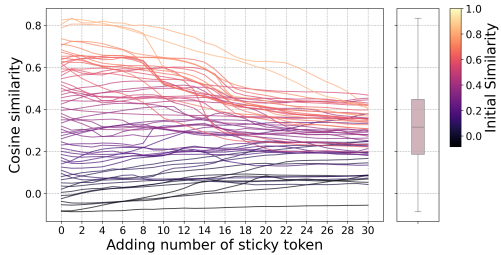

Figure 5: gte-base-en-v1.5 + token: 一

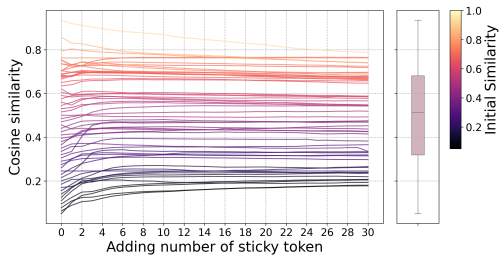

Figure 6: bge-base-en-v1.5 + token: www

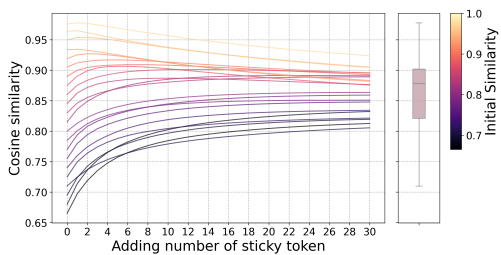

Figure 7: instructor-base + token: lucrarea

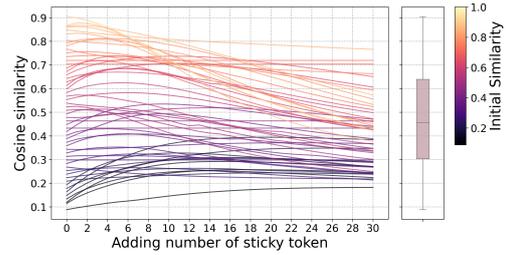

Figure 8: sup-simcse-bert-base-uncased + token: [SEP]

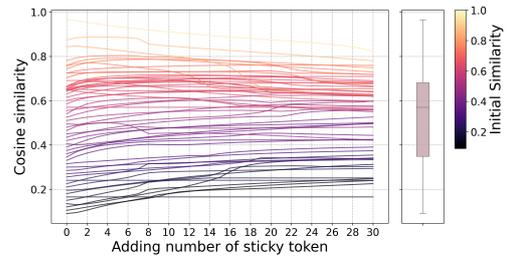

Figure 9: UAE-Large-V1 + token: [SEP]

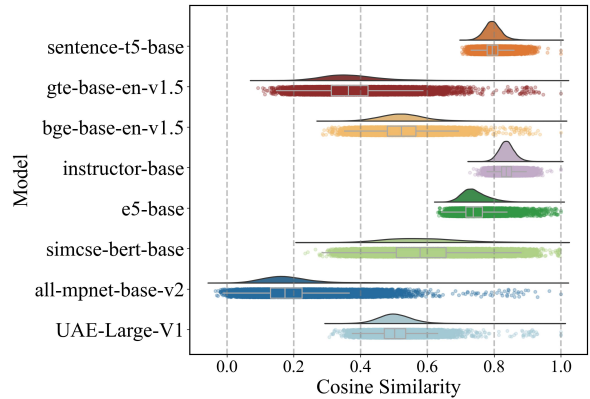

Figure 10: Similarity distribution for different text embedding models' vocabulary tokens. We use token embeddings as a surrogate for text embeddings since both share the same embedding space.

The results of the phenomenon are shown in Figure 6, 5, 8, 9, and 7, and for more examples, please refer to our repository.

## B Alternative Insertion Operations: Deletion or Replacement

We first explain the insertion operations in Section 3.2 in detail. As illustrated in Figure 11, inserting a token $t$ into a sentence $s$ can happen in different ways, including (1) repeatedly adding $t$ at the beginning (prefix), (2) repeatedly adding $t$ at the end (suffix), or (3) adding $t$ at random positions. Real-world scenarios might involve more sophisticated insertion strategies. Here we discuss why deletion or replacement operations are not considered in our work.

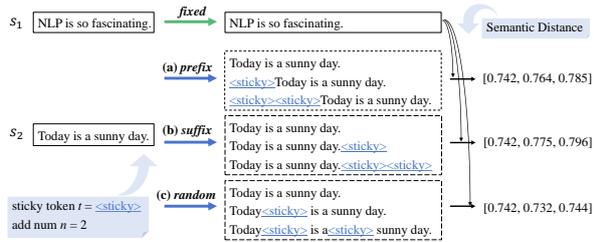

Figure 11: Inserting operations of token into sentence.

**Deletion** Deletion is the inverse operation of addition. Deleting a token in a sentence is the reverse process of adding a token to a sentence. Since sticky tokens are relatively rare, it is challenging to gather a sufficient number of sentences that naturally contain them.

**Replacement** Replacing a token in a sentence can be viewed as first deleting the original token and then inserting a new one, which modifies the sentence's semantic in two steps. As shown in Figure 12, the experimental setup remains consistent with Section 3.1. When tokens are introduced through replacement, the shift in sentence similarity appears less gradual compared to the smoother pattern observed in Figure 2b. This suggests that, unlike addition, replacement leads to larger and less granular semantic changes, making it unsuitable as the most basic unit of semantic changes.

For simplicity in modeling and broader applicability, we exclude deletion and replacement operations from the definition of the insertion operations $\mathbb{I}$ in Section 3.2.

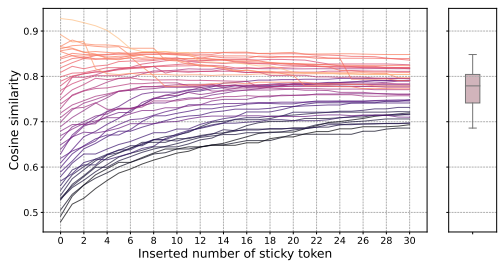

Figure 12: Effect of token replacement operation on sentence similarity. This figure illustrates the impact of replacing tokens in sentences with the sticky token lucrarea on sentence similarity, as measured using the ST5-base model. Sentence pairs were randomly selected from Wikipedia, and their similarity was calculated both before and after the replacement of multiple lucrarea tokens.

## C Conjecture of Explanation: Anisotropic Embedding Space Makes Sticky Token Possible

We first provide partial background knowledge about the spatial properties of context embedding space, and then propose a conjecture for the potential reason why the sticky token exists in text embedding models.

**Anisotropic text embedding space** Isotropy refers to the property that embeddings are uniformly distributed around the origin. Previous studies (Wang et al., 2019; Arora et al., 2017; Fuster Baggetto and Fresno, 2022) demonstrate that Transformer-based models typically produce anisotropic embedding spaces. The geometric interpretation of anisotropy is that the word representations all occupy a narrow cone in the vector space rather than being uniform in all directions; the greater the anisotropy, the narrower this cone (Mimno and Thompson, 2017; Ethayarajh, 2019). This phenomenon has been empirically observed in pre-trained Transformers like BERT and GPT-2 (Machina and Mercer, 2024).

We also construct a simple empirical experiment to demonstrate the anisotropic context embedding space of mainstream text embedding models. we use word embeddings as a surrogate since words and contexts share the same embedding space (Gao et al., 2019; Fuster Baggetto and Fresno, 2022). If the word embeddings exhibits some misleading properties, the context embeddings will also be problematic, and vice versa.

We first extract the *vocabulary* of the model, then take each token in the dictionary as a separate sentence and gets its embeddings. More specifically, the embeddings for tokens are computed after they have passed through all transformer layers (i.e., using the final layer's output). Finally we compute the pairwise similarity between embeddings, and the results are presented in Figure 10. For more models, the mean values and standard deviations of cosine similarity across vocabulary embeddings are presented in Table 3.

| Model | Mean | Std | Model | Mean | Std |
|---|---|---|---|---|---|
| all-MiniLM-L6-v2 | 0.1998 | 0.1068 | gtr-t5-xl | 0.4824 | 0.0562 |
| all-mpnet-base-v2 | 0.1876 | 0.0885 | gtr-t5-xxl | 0.4774 | 0.0543 |
| bge-base-en-v1.5 | 0.5254 | 0.0673 | instructor-base | 0.8373 | 0.0234 |
| bge-large-en-v1.5 | 0.5716 | 0.0482 | instructor-large | 0.8144 | 0.0229 |
| bge-small-en-v1.5 | 0.5694 | 0.0602 | instructor-xl | 0.5544 | 0.0488 |
| e5-base | 0.7430 | 0.0403 | nomic-embed-text-v1 | 0.3360 | 0.0630 |
| e5-large | 0.7311 | 0.0351 | nomic-embed-text-v1.5 | 0.4167 | 0.0610 |
| e5-mistral-7b-instruct | 0.7354 | 0.0579 | sentence-t5-base | 0.7959 | 0.0261 |
| e5-small | 0.8306 | 0.0392 | sentence-t5-large | 0.7634 | 0.0281 |
| GritLM-7B | 0.6271 | 0.1838 | sentence-t5-xl | 0.7167 | 0.0341 |
| gte-base | 0.7647 | 0.0256 | sentence-t5-xxl | 0.7362 | 0.0310 |
| gte-base-en-v1.5 | 0.3730 | 0.0892 | SFR-Embedding-2_R | 0.7264 | 0.0638 |
| gte-large | 0.7788 | 0.0218 | SFR-Embedding-Mistral | 0.6806 | 0.0598 |
| gte-large-en-v1.5 | 0.5390 | 0.0651 | sup-simcse-bert-base-uncased | 0.5866 | 0.1110 |
| gte-Qwen2-1.5B-instruct | 0.3510 | 0.2746 | sup-simcse-bert-large-uncased | 0.4512 | 0.1081 |
| gte-Qwen2-7B-instruct | 0.2594 | 0.2477 | sup-simcse-roberta-base | 0.8783 | 0.0361 |
| gte-small | 0.7874 | 0.0225 | sup-simcse-roberta-large | 0.4995 | 0.1039 |
| gtr-t5-base | 0.5155 | 0.0548 | UAE-Large-V1 | 0.5052 | 0.0523 |
| gtr-t5-large | 0.5577 | 0.0451 | jina-embeddings-v3 | 0.5281 | 0.06095 |
| | | | KaLM-v1 | 0.8565 | 0.0389 |
| | | | KaLM-v1.5 | 0.8523 | 0.0360 |

Table 3: The average cosine similarity values (**Mean**) and their standard deviations (**Std**) for token embeddings (of vocabulary) across various models.

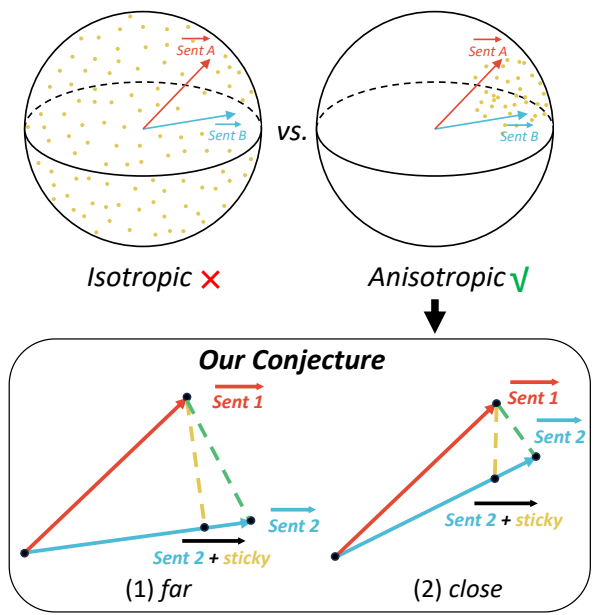

Figure 13: Our conjecture about sticky tokens, based on the anisotropy of the embedding space.

Previous research has demonstrated that if word representations are isotropic (i.e., directionally uniform), then the average cosine similarity between words would be 0 (Arora et al., 2017; Ethayarajh, 2019). The closer this average is to 1, the more anisotropic the representations. As illustrated in Figure 10, we observe that the similarity distributions for most models follow a Gaussian distribution with a non-zero mean, indicating that these models exhibit anisotropic embedding spaces. Additionally, it is noteworthy that the *mean* of the ST5-Base model's similarity distribution is very close to the sentences similarities' *median* value of 0.8, as depicted in Figure 2b in Section 3.1. This suggests that the sticky token is likely pulling sentence pairs toward a dominant direction in the embedding space. Based on the above observations, we propose a **conjecture** to explain the existence of sticky tokens.

**Conjecture** As illustrated in Figure 13, the anisotropy of the model embedding space, indicating that word representations occupy narrow cone-shaped regions in vector space. Sticky tokens tend to pull a sentence toward a specific focal point in the embedding space, potentially the origin. (1) If the sentences are sufficiently far apart, the new distance (yellow) is more likely to be shorter than the original distance (green). (2) However, if the sentences are already very close to each other, this may negatively impact performance.

Please note that *these are merely some of our conjectures, and rigorous validation will be required in the future*.

## D Methodology Details

Based on Definition 1 in Section 3.2, we provide a detailed description of our proposed method, STD, which is designed to effectively detect sticky tokens in text embedding models. As shown in Figure 3, our method takes a target text embedding model and a string dataset as inputs, then reports its sticky tokens of its vocabulary.

**Motivation** For the detection, Definition 1 suggests to track sentence pairwise-similarity changes across any pairs for any tokens in a vocabulary, which can be computationally expensive. Figure 2 suggests that actually the influence on just a portion of sentence pairs may be sufficient to differentiate sticky tokens from normal ones. For instance, they obviously and efficiently increase similarity of sentence pairs towards the *mean* of token pairwise-similarity distribution (whose similarity initially is below that *mean*). From this, we employ an efficient detection procedure which first filter sentences to track and then shortlist candidate tokens. Specifically:

1. Sentence pair filter, filter out sentence pairs with initial similarity above the *mean* of the distribution 1.

2. Token filter, filter out those undecodable or unreachable tokens.

3. Shortlisting, shortlist tokens via sticky scoring.

4. Validation, validate whether the shortlisted tokens are indeed sticky ones based on Definition 1.

### D.1 Sentence Pair Filtering

We observed in Figure 2 that compared with normal ones, sticky tokens tend to obviously bring closer those sentences whose initial similarity is below the mean $u$ of the initial pairwise-similarity distribution of tokens.

Formally, given a model $E$, the *mean* of the pairwise semantic distance between (embeddings of) its tokens can be computed as:

$$u = \frac{2}{|\mathcal{V}|(|\mathcal{V}|-1)} \sum_{i=1}^{|\mathcal{V}|-1} \sum_{j=i+1}^{|\mathcal{V}|} Sim(t_i, t_j)$$

where $\mathcal{V}$ denotes the model's vocabulary, and $t_i, t_j \in \mathcal{V}$ represent distinct tokens.

We can choose such sentence pairs from $\mathbb{S}$ to check whether the influence of a token aligns with that of sticky tokens via multiple insertion. We denote the set formed by filtered sentence pairs as:

$$\mathbb{P}_f = \{(s_1, s_2) \mid Sim(s_1, s_2) < u, s_1, s_2 \in \mathbb{S}\}$$

### D.2 Token Filtering

The overall process of our token filtering stage is shown in Table 4. The core idea of the token filter module is to classify each token by decoding and then re-encoding it, ensuring it meets specific classification criteria. Specifically, if a tokenizer for one model adds spaces to the start of tokens or applies default reprocessing, we prepend a special prefix "«" to each token to ensure consistent encoding and decoding. Then, we filter out tokens based on the following categories:

- *Undecodable*: Tokens that cannot be decoded, usually containing illegal characters. These tokens are usually the result of partial UTF-8 sequences, where a sequence of bytes cannot be properly converted into a Unicode character, due to containing only part of a UTF encoding for a character. This is typical for "fallback byte" tokens in the 0x80-0xFF range, can also include tokens with other partial Unicode characters.

- *Unreachable*: Tokens that cannot be restored to their original token ID through the decoding and re-encoding process, which means they are never the result of tokenizing text. Such tokens are typically the result of tokenizer configuration errors or conflicts between trained and manually added vocabulary. Since this test does not work when tokens can not be decoded to a string, we exclude undecodable tokens from this category.

- *Special*: Tokens that are manually predefined symbols used to represent specific meanings or control the model's behavior, such as [CLS], [SEP], </s>, etc. We identify special tokens using the patterns <...> and [...] and list them separately from unreachable tokens.

- Tokens not in any of the other categories, which constitute the vast majority.

**Classification Criteria**

Let $D : \mathbb{N} \to \Sigma^*$ be the tokenizer's decoding function and $E : \Sigma^* \to \mathbb{N}$ its encoding function, where $\Sigma$ is the Unicode character set. For a token ID $x$, :

**Undecodeable**
$x \in \mathcal{U} \iff D(x)$ throws decoding error
Where illegal UTF-8 sequences satisfy:
$\exists b_i \in \text{bytes}(D(x)) \ s.t. \ \neg\text{ValidUTF-8}(b_{1:n})$

**Unreachable**
$x \in \mathcal{R} \iff D(x) \text{ succeeds } \land \ E(D(x)) \neq x$

**Special Tokens**
$x \in \mathcal{S} \iff D(x)$ matches patterns $\langle...\rangle$ or $[...]$

**Token Filtering Pipeline**
$ValidTokens \ \mathcal{V}^* = \{x \mid x \notin (\mathcal{U} \cup \mathcal{R})\}$

Table 4: Formalizing token classification criteria and token filtering pipeline.

During the classification process, we first decode each token ID to string. If decoding fails, the token is classified as *undecodable*. Next, we encode the decoded string and check if it can be restored to the original token ID. If it cannot, the token is classified as *unreachable*. If it meets the characteristics of a special token, it is classified as *special*. We filter out *undecodable* and *unreachable* tokens from our sticky token detection pipeline.

The *valid* UTF-8 characters in Table 4 can be summarized as follows:

- 1-byte: $b_1 \in [0x00, \ 0x7F]$
- 2-byte: $b_1 \in [0xC2, \ 0xDF], b_2 \in [0x80, \ 0xBF]$
- 3-byte: $b_1 \in [0xE0, \ 0xEF], b_{2:3} \in [0x80, \ 0xBF]$
- 4-byte: $b_1 \in [0xF0, \ 0xF4], b_{2:4} \in [0x80, \ 0xBF]$

*Undecodable* tokens violate these byte constraints.

### D.3 Shortlisting with Sticky Scoring

The previous two steps help us reduce the searching cost of sticky tokens to some extent. Faced with the filtered tokens and sentence pairs, a straightforward way to judge which ones in $\mathcal{V}^*$ are sticky is to track their influence on arbitrary sentence pairs in $\mathbb{P}_f$. This can still be time-consuming and we opt to track on some sparsely sampled pairs first, to shortlist tokens. The core consideration is how to measure whether the influence a token brings via insertion to the sampled sentence pairs aligns with our expectation[12]. Below we introduce sticky score.

Denote the $k$ sampled sentence pairs as: $p_j \in \mathbb{S}_f, p_j = (s_1^j, s_2^j)$. Let $\Delta_{t,\mathcal{I},p}^j = Sim(s_1^j, s_2^{j'}) - Sim(s_1^j, s_2^j)$ denote the change in similarity between $s_1^j, s_2^j$ after inserting $t$. For each pair, one of the sentence gets inserted[13] with token $t$ via operation $\mathcal{I}$. For example, for $s_1^j, s_2^j$, let $\Delta_{t,\mathcal{I},p}^j = Sim(s_1^j, s_2^{j'}) - Sim(s_1^j, s_2^j)$ denote the change in their similarity. For all the pairs, denote the change as $\mathcal{L}_{t,f,p} = \left[\Delta_{t,f,p}^1, \Delta_{t,f,p}^2, \ldots, \Delta_{t,f,p}^k\right] \in \mathbb{R}^k$.

We measure the influence of token insertion from two key aspects: the magnitude and frequency of changes in directional similarity. Let $M_{(t,f,p)}^+ = \sum_{i=1}^k \max(\Delta^j, 0)$ ($M_{(t,f,p)}^- = \sum_{j=1}^k |\min(\Delta^i, 0)|$) denote the cumulative amount of similarity increase (decrease), and $F_{(t,f,p)}^+ = \frac{1}{k}\sum_{j=1}^k \mathbb{I}(\Delta^{(i)}>0)$ ($F_{(t,f,p)}^- = \frac{1}{k}\sum_{j=1}^k \mathbb{I}(\Delta^{(j)}<0)$) denote the frequency of observing similarity increase (decrease) in $\mathcal{L}_{(t,f,p)}$. $\mathbb{I}(\cdot)$ is an indicator function which takes 1 if $(\cdot)$ is true.

By integrating the above influence measure, we propose sticky score:

$$SS_{\mathcal{I},p}(t) = \frac{M^+ + \alpha F^+ +}{M^- + \beta F^- + \gamma + Sim(s_1, t)}$$

where $M^+ + \alpha F^+$ rewards positive values (i.e., increasing similarity) in $\mathcal{L}$, and $M^- + \beta F^-$ penalizes negative values. $Sim(s_1, t)$ is used to penalizes any semantic proximity between token $t$ and the target sentence $s_1$, preventing artificially inflated anomaly scores when their meanings are closely aligned. $\gamma > 0$ is a small constant (e.g., $\gamma = 10^{-8}$) to ensure numerical stability. Parameters $\alpha$

---

[12]Recall for $(s_1, s_2) \in \mathbb{P}_f$, $Sim(s_1, s_2) < u$, and a sticky token should make $|Sim(s_1, \mathcal{I}(s_2, t, k)) - u|$ smaller.

[13]As $s_1/s_2$ is randomly chosen from $\mathbb{S}_f$, their order does not matter and w.lo.g the insertion is for $s_2$.

---

**Algorithm 1:** Token Validation

**Input:** $\mathcal{C}$: the set of candidate tokens,
$\mathbb{P}_f$: the set of filtered sentence pairs,
$\mathbb{I}$: the set of insertion operations,
$E$: embedding model, $n$: insertion number,
$u$: mean similarity of vocab token embeddings, $\epsilon$: threshold
**Output:** $\Omega$: validated sticky tokens

1 $\Omega \leftarrow \emptyset$
2 **forall** $t \in \mathcal{C}$ **do**
3     $is\_sticky \leftarrow$ True;
4     **forall** $(s_1, s_2) \in \mathbb{P}_f$ **do**
5        **for** $\mathcal{I} \in \mathbb{I}$ **do**
6           $s_2^* \leftarrow \mathcal{I}(s_2, t, n); e_1 \leftarrow E(s_1)$,
           $e_2^* \leftarrow E(s_2^*)$;
7           **if** $|Sim(e_1, e_2^*) - u| > \epsilon$ **then**
8              $is\_sticky \leftarrow$ False;
9              **break**
10        **if** $\neg is\_sticky$ **then**
11           **break**;
12     **if** $is\_sticky$ **then**
13        $\Omega \leftarrow \Omega \cup \{t\}$
14 **return** $\Omega$;

---

and $\beta$ are tuning factors that allow the detection to balance the consideration of magnitude and frequency factors.

By aggregating the influence introduced by all types of insert operations, across all the filter sentence pairs, we obtain an overall sticky metric for token $t$:

$$SS(t) = \sum_{\mathcal{I} \in \mathbb{I}} \sum_{p \in \mathbb{P}} SS_{\mathcal{I},p}(t)$$

$SS(t)$ measures how well (the influence of) token $t$ fits our expectation or what characterizes sticky tokens. The higher the value of $SS(t)$, the more likely $t$ is a sticky token. Given an embedding model, we rank all its tokens based on their values of $SS(t)$ and shortlist those ranked top 2% to obtain our candidate token set $\mathcal{C}$. Note that only a sampled set of sentence pairs are used in calculating $SS(t)$ and we need to further validate whether the shortlisted tokens in $\mathcal{C}$ are indeed sticky ones.

### D.4 Validation

We validate whether the previously shortlisted tokens are indeed sticky ones by determining

**Algorithm 2:** Adaptive Threshold

**Input:** $E$: target embedding model, $G_E$: the set of $G_E(t)$ values for all tokens $t \in \mathcal{C}$, $\alpha$: the hyperparameter (default: 1.5)

**Output:** $\epsilon$: the threshold for model $E$

1 $\epsilon \leftarrow \emptyset$;
2 Calculate quartiles:
$Q1^E \leftarrow \text{quantile}(G_E, 0.25)$;
$Q3^E \leftarrow \text{quantile}(G_E, 0.75)$;
$IQR^E \leftarrow Q3^E - Q1^E$;
3 Compute threshold:
$\epsilon^E \leftarrow Q3^E + \alpha \times IQR^E$;
4 **return** $\epsilon$;

whether it adheres to the definition of a sticky token (Definition 1). At this stage, we use all the samples in the set of sentence pairs $\mathbb{P}_f$ from Section D.1. The overall process of validation stage is shown in Algorithm 1.

### D.4.1 Adaptive Threshold

As mentioned above, for the tokens in the candidate token set: $t \in \mathcal{C}$, we need to calculate $|Sim(s_1, \mathcal{I}(s_2, t, k)) - u|$ according to Definition 1, and for clarity we denote this value as $G_E(t)$ for token $t$ and model $E$:

$$G_E(t) = |Sim(s_1, \mathcal{I}(s_2, t, k)) - u|,$$

$$G_E = \{G_E(t) \mid t \in \mathcal{C}\}$$

where $G_E$ represents the set of $G_E(t)$ values for all tokens $t$ in the candidate token set $\mathcal{C}$ of the model $E$. As detailed in Algorithm 2, we present an adaptive thresholding algorithm inspired by statistical anomaly detection theory. This approach leverages the interquartile range (IQR) to dynamically identify outliers without being constrained by distributional differences between models (Grubbs, 1969).

### E Detection Experiment Details

**Dateset** As mentioned in Section 4.1, the detection of sticky tokens needs sentences to feed into embedding models and computing the semantic similarity between embeddings. The NLP task most closely related to this process is Semantic Textual Similarity (STS). For our analysis, we utilize the STS datasets included in the widely recognized Massive Text Embedding Benchmark (MTEB) (Muennighoff et al., 2023), which includes STS12, STS13, STS14, STS15, STS16, STS17, STS22, STSBenchmark, BIOSSES, SICK-R [14] (Agirre et al., 2012, 2013, 2014, 2015, 2016). We used the test sets of these datasets, each containing between 1k and 20k sentences. Most datasets are monolingual English, and for multilingual datasets, we only used their English subsets.

**Target Text Embedding Models** As illustrated in Table 5, we evaluated STD using models from 14 different model families. For embedding models that support Matryoshka Representation Learning (Kusupati et al., 2024), we utilize the highest-dimensional vectors with default parameters. For models that require prompts, we employ the default prompts as specified in the original papers.

**Ablation Study** Following Definition 1 and Section 4.3, we need to choose values for $n$ (the number of insertions), $k$ (the number of sampled sentence-pairs), and the threshold $\epsilon$ for model validation. We conducted some ablation studies to balance between computational efficiency and detection effectiveness.

First of all, we need to establish a certain understanding of the running time of text embedding models. Our code has implemented the batch data parallelism. For a 7B embedding model, when $n = 10$, sentence pair $k = 10$, and the number of sentence pairs $|\mathbb{S}| = 200$, it will take 25 hours for the detection pipline to detecting the all vocabulary, so it is impractical to exceed this configured parameter number. Therefore, we define this set of parameters as the *upper bound*, and the results obtained under this configuration serve as the *ground truth* for sticky token detection. Additionally, we conducted an ablation study on the ST5-base model, with the results presented in Table 6. To balance computational efficiency and detection effectiveness, we selected $n = 8$ and $k = 5$.

The corresponding thresholds used in our work for each model are provided in Table 7. We obtain this set of parameters by using Algorithm 2. Please note that threshold values in Table 7 are computed for arbitrary sentence pairs, and values such as 0.19 are expected given the varying similarity distributions across different embedding models.

---
[14] https://huggingface.co/mteb?search_datasets=sts#:~:text=2-,Datasets,-13

| Model Family | Model Names |
|---|---|
| Sentence-BERT (Reimers and Gurevych, 2019b) | all-MiniLM-L6-v2, all-mpnet-base-v2 |
| SimCSE (Gao et al., 2022) | sup-simcse-bert-base-uncased, sup-simcse-bert-large-uncased, sup-simcse-roberta-base, sup-simcse-roberta-large |
| Sentence-T5 (Ni et al., 2021a) | sentence-t5-base, sentence-t5-large, sentence-t5-xl, sentence-t5-xxl |
| GTR (Ni et al., 2021b) | gtr-t5-base, gtr-t5-large, gtr-t5-xl, gtr-t5-xxl |
| Instructor (Su et al., 2023) | instructor-base, instructor-large, instructor-xl |
| E5 (Wang et al., 2024b,c) | e5-small, e5-base, e5-large, e5-mistral-7b-instruct |
| BGE (Xiao et al., 2024) | bge-small-en-v1.5, bge-base-en-v1.5, bge-large-en-v1.5 |
| AnglE (Li and Li, 2024) | UAE-Large-V1 |
| Nomic (Nussbaum et al., 2025) | nomic-embed-text-v1, nomic-embed-text-v1.5 |
| GTE (Li et al., 2023) | gte-small, gte-base, gte-large, gte-base-en-v1.5, gte-large-en-v1.5, gte-Qwen2-1.5B-instruct, gte-Qwen2-7B-instruct |
| jina (Sturua et al., 2024) | jina-embeddings-v3 |
| KaLM (Hu et al., 2025) | KaLM-embedding-multilingual-mini-instruct-v1 (abbreviated as KaLM-instruct-v1), KaLM-embedding-multilingual-mini-instruct-v1.5（abbreviated as KaLM-instruct-v1.5） |
| GritLM (Muennighoff et al., 2024) | GritLM-7B |
| SFR (Yavuz et al., 2024) | SFR-Embedding-2_R, SFR-Embedding-Mistral |

Table 5: Target text embedding models used in the experiments.

| Set(n, k) | Runtime (h) | Accuracy (%) | F1-Score |
|---|---|---|---|
| (5, 3) | 1.1 | 83.7 | 0.812 |
| (6, 4) | 1.8 | 88.4 | 0.862 |
| (7, 5) | 2.1 | 90.6 | 0.891 |
| **(8, 5)** | **2.5** | **92.1** | **0.907** |
| (9, 6) | 3.3 | 93.8 | 0.923 |
| (10, 10) | 4.9 | 100.0 | 1.000 |
| (8, 6) | 2.7 | 91.2 | 0.896 |
| (7, 4) | 1.9 | 89.1 | 0.878 |

Table 6: Ablation study on parameter selection for sticky token detection

| Model | Threshold | Model | Threshold |
|---|---|---|---|
| all-MiniLM-L6-v2 | 0.0865 | gtr-t5-large | 0.0451 |
| all-mpnet-base-v2 | 0.0742 | gtr-t5-xl | 0.0562 |
| bge-base-en-v1.5 | 0.1649 | gtr-t5-xxl | 0.0543 |
| bge-large-en-v1.5 | 0.1686 | instructor-base | 0.0690 |
| bge-small-en-v1.5 | 0.1596 | instructor-large | 0.0706 |
| e5-base | 0.0819 | instructor-xl | 0.1165 |
| e5-large | 0.0796 | nomic-embed-text-v1 | 0.0362 |
| e5-mistral-7b-instruct | 0.1254 | nomic-embed-text-v1.5 | 0.0254 |
| e5-small | 0.0777 | sentence-t5-base | 0.1106 |
| GritLM-7B | 0.2089 | sentence-t5-large | 0.1153 |
| gte-base | 0.0546 | sentence-t5-xl | 0.1303 |
| gte-base-en-v1.5 | 0.0892 | sentence-t5-xxl | 0.1233 |
| gte-large | 0.0652 | SFR-Embedding-2_R | 0.1243 |
| gte-large-en-v1.5 | 0.0651 | SFR-Embedding-Mistral | 0.0568 |
| gte-Qwen2-1.5B-instruct | 0.1841 | sup-simcse-bert-base-uncased | 0.1832 |
| gte-Qwen2-7B-instruct | 0.1542 | sup-simcse-bert-large-uncased | 0.1952 |
| gte-small | 0.0542 | sup-simcse-roberta-base | 0.1523 |
| gtr-t5-base | 0.0548 | sup-simcse-roberta-large | 0.1644 |
| UAE-Large-V1 | 0.1721 | jina-embeddings-v3 | 0.1281 |
|  |  | KaLM-v1 | 0.0764 |
|  |  | KaLM-v1.5 | 0.0925 |

Table 7: Threshold to validate sticky token. We obtain this set of parameters by using Algorithm 2. Note that these values are derived from the *standard deviation*, not the *variance*, between sentence similarities.

| Category | Task |
|---|---|
| Retrieval | SciFact<br>ArguAna<br>NFCorpus |
| Reranking | SciDocsRR<br>StackOverflowDupQuestions |
| Clustering | BiorxivClusteringS2S<br>MedrxivClusteringS2S<br>TwentyNewsgroupsClustering |
| Pair Classification | SprintDuplicateQuestions |
| Classification | Banking77Classification<br>EmotionClassification<br>MassiveIntentClassification |
| STS | STS16<br>SICK-R<br>STSBenchmark |
| Summarization | SummEval |

Table 8: The subset of MTEB evaluation benchmark used in downstream impact studies.

# F  Downstream task detail

We assess how sticky tokens degrade contextual representations through sequence-level evaluation on text embedding tasks.

**Set up** we evaluate on the Massive Text Embedding Benchmark (MTEB) (Muennighoff et al., 2023), a collection of 7 diverse embedding task categories. MTEB consists of diverse small and large embedding tasks. To speed up the evaluation, we consider a representative subset of 16 tasks from MTEB for our analyses, presented in Table 8[15]. To make sure that our analyses are not biased towards one specific category or task, this subset includes tasks from each category with almost the same proportion compared to the full MTEB.

| Categories → | Classification | | | Clustering | | | Pair Classification | Reranking | | Retrieval | | | STS | | | Summarization |
|---|---|---|---|---|---|---|---|---|---|---|---|---|---|---|---|---|
| Datasets → | Banking77 | Emotion | MassiveIntent | Biorxiv | Medrxiv | TwentyNews groups | SprintDuplicate Questions | StackOverflow DupQuestions | SciDocsRR | SciFact | ArguAna | NFCorpus | SICK-R | STS16 | STS Benchmark | SummEval |
| sentence-t5-base | 76.60 | 51.34 | 69.70 | 23.11 | 26.03 | 49.27 | 91.23 | 48.46 | 73.96 | 45.76 | 44.84 | 28.64 | 80.18 | 84.03 | 85.52 | 31.39 |
| w/ normal token | 75.73 | 51.30 | 66.57 | 20.04 | 25.06 | 37.17 | 87.86 | 44.85 | 72.05 | 44.58 | 45.41 | 28.48 | 76.72 | 79.69 | 81.32 | 30.32 |
| w/ sticky oken | 75.20 | 50.20 | 66.83 | **15.02** | **20.41** | 35.38 | 88.39 | 45.16 | 71.17 | **26.76** | 42.14 | **13.65** | 76.32 | 79.26 | 81.24 | 30.84 |
| gte-base-en-v1.5 | 86.72 | 46.34 | 77.67 | 37.39 | 32.31 | 48.66 | 95.03 | 52.18 | 85.16 | 76.79 | 63.65 | 35.85 | 79.38 | 85.02 | 86.06 | 31.35 |
| w/ normal token | 85.87 | 46.10 | 74.92 | 36.31 | 32.01 | 44.68 | 94.19 | 50.00 | 84.67 | 73.36 | 62.14 | 35.22 | 77.36 | 81.75 | 83.65 | 31.87 |
| w/ sticky token | 84.44 | 44.26 | 70.36 | 36.11 | 31.03 | 45.20 | 89.97 | 46.16 | 83.77 | 75.41 | 61.58 | 35.77 | 74.85 | 76.96 | 78.49 | 30.46 |
| bge-base-en-v1.5 | 83.99 | 54.61 | 72.64 | 36.62 | 31.68 | 50.75 | 96.37 | 54.62 | 87.49 | 73.76 | 63.62 | 36.81 | 80.30 | 85.47 | 86.42 | 31.04 |
| w/ normal token | 82.57 | 52.70 | 66.98 | 36.20 | 30.74 | 44.27 | 95.18 | 50.94 | 86.59 | 72.91 | 60.63 | 37.15 | 76.10 | 80.97 | 82.02 | 29.97 |
| w/ sticky oken | 82.31 | 51.98 | 67.62 | 35.93 | 31.06 | 43.36 | 94.95 | 50.99 | 86.61 | 73.70 | 61.31 | 37.05 | 77.80 | 80.11 | 81.72 | 30.31 |
| instructor-base | 76.92 | 48.48 | 66.00 | 26.40 | 28.38 | 52.77 | 92.06 | 50.66 | 79.36 | 57.88 | 51.18 | 30.76 | 80.02 | 84.78 | 85.85 | 30.57 |
| w/ normal token | 75.07 | 45.79 | 62.38 | 18.05 | 23.13 | 50.64 | 88.39 | 47.66 | 77.92 | 57.70 | 47.45 | 29.77 | 75.48 | 77.97 | 79.99 | 30.37 |
| w/ sticky oken | 76.37 | 47.66 | 64.62 | 26.05 | 26.55 | 50.55 | 91.30 | 49.67 | 76.63 | **43.47** | 47.03 | **23.11** | 78.86 | 81.96 | 84.21 | 29.17 |
| e5-base | 76.27 | 51.85 | 66.65 | 29.92 | 27.67 | 43.75 | 94.19 | 48.18 | 81.01 | 71.88 | 53.03 | 37.09 | 80.66 | 84.49 | 86.35 | 31.04 |
| w/ normal token | 74.85 | 49.91 | 63.00 | 28.94 | 26.51 | 22.15 | 91.37 | 44.11 | 79.85 | 71.36 | 51.13 | 37.15 | 76.01 | 78.17 | 79.42 | 30.76 |
| w/ sticky oken | 75.13 | 49.30 | 61.91 | 27.02 | 24.92 | 20.00 | 91.53 | 44.80 | 80.03 | 70.95 | 49.14 | 37.01 | 77.17 | 77.68 | 80.19 | 29.99 |
| simcse-bert-base | 75.49 | 45.69 | 67.21 | 25.70 | 25.85 | 31.67 | 81.74 | 40.32 | 71.14 | 33.89 | 39.56 | 13.49 | 80.62 | 80.71 | 82.69 | 31.17 |
| w/ normal token | 71.42 | 43.49 | 60.38 | 25.11 | 25.19 | 28.40 | 76.54 | 37.34 | 70.02 | 33.66 | 36.79 | 13.45 | 77.53 | 75.82 | 78.32 | 30.76 |
| w/ sticky oken | 72.40 | 43.34 | 61.03 | 24.80 | 25.17 | 29.22 | 76.51 | 38.31 | 70.25 | **29.89** | 38.38 | **8.84** | 77.74 | 77.05 | 79.53 | 30.18 |
| all-mpnet-base-v2 | 81.7 | 42.23 | 69.76 | 34.82 | 33.42 | 50.07 | 90.15 | 51.98 | 88.65 | 65.57 | 46.52 | 33.29 | 80.59 | 80.03 | 83.42 | 27.49 |
| w/ normal token | 79.7 | 40.01 | 64.1 | 33.93 | 32.55 | 39.2 | 86.24 | 47.33 | 87.77 | 65.14 | 44.25 | 33.2 | 77.8 | 68.13 | 74.11 | 28.3 |
| w/ sticky oken | 79.64 | 40.65 | 65.02 | 34.05 | 32.16 | 39.28 | 85.87 | 47.79 | 87.77 | 64.81 | 43.98 | 33.16 | 78.04 | 68.2 | 73.19 | 26.17 |
| UAE-Large-V1 | 87.73 | 51.72 | 76.24 | 37.24 | 31.18 | 51.72 | 97.24 | 55.32 | 87.49 | 73.91 | 66.15 | 37.61 | 82.62 | 86.61 | 89.06 | 32.03 |
| w/ normal token | 86.09 | 48.16 | 72.13 | 35.79 | 30.96 | 40.48 | 96.23 | 50.44 | 86.75 | 74.51 | 63.67 | 37.70 | 80.72 | 80.43 | 84.23 | 31.99 |
| w/ sticky oken | 86.56 | 50.43 | 72.79 | 35.98 | 30.94 | 47.20 | 96.52 | 52.44 | 86.94 | **72.63** | 63.48 | 37.79 | 81.53 | 83.13 | 86.00 | 30.84 |
| e5-mistral-7b-instruct | 78.60 | 48.41 | 71.15 | 34.47 | 32.29 | 47.31 | 89.88 | 46.56 | 82.09 | 75.18 | 53.88 | 33.26 | 80.76 | 84.83 | 84.59 | 31.07 |
| w/ benign token | 77.56 | 46.93 | 68.73 | 31.85 | 30.32 | 44.84 | 89.10 | 45.15 | 80.77 | 74.71 | 54.18 | 34.57 | 79.13 | 79.30 | 81.27 | 30.15 |
| w/ sticky oken | 74.95 | 40.16 | 65.32 | 28.92 | 28.37 | 40.78 | 80.57 | 41.01 | 79.07 | 72.21 | 55.30 | 33.17 | 74.85 | 74.08 | 67.46 | 27.15 |
| gte-Qwen2-7B-instruct | 84.00 | 55.28 | 77.46 | 39.16 | 33.34 | 52.34 | 93.13 | 52.87 | 86.25 | 79.55 | 64.71 | 40.33 | 78.06 | 82.82 | 81.61 | 30.46 |
| w/ benign token | 82.89 | 54.43 | 74.06 | 38.05 | 32.47 | 48.35 | 93.08 | 49.39 | 85.66 | 79.68 | 63.66 | 40.59 | 71.79 | 74.24 | 72.30 | 29.95 |
| w/ sticky oken | 81.48 | 53.22 | 72.84 | 32.77 | 29.64 | 47.50 | 86.69 | 43.40 | 81.49 | 72.61 | 55.03 | 34.45 | 62.75 | 71.69 | 66.81 | 28.38 |
| GritLM-7B | 70.44 | 36.03 | 62.90 | 23.67 | 24.13 | 19.57 | 58.58 | 35.56 | 60.05 | 44.57 | 37.33 | 6.99 | 58.07 | 57.60 | 48.32 | 24.11 |
| w/ benign token | 62.05 | 35.13 | 56.75 | 14.75 | 18.80 | 10.94 | 44.45 | 29.86 | 53.90 | 39.52 | 33.61 | 6.25 | 27.69 | 32.07 | 27.29 | 25.80 |
| w/ sticky oken | 56.42 | 34.91 | 49.36 | 19.41 | 15.82 | 10.04 | 40.73 | 26.18 | 48.93 | 36.41 | 29.73 | 6.19 | 26.64 | 41.04 | 23.69 | 22.81 |

Table 9: Results on downstream tasks. We compared the performance of 11 models, comparing their baseline results with perturbation of sticky tokens and normal tokens.

For each model under investigation, we have previously identified its associated list of sticky tokens, as delineated in Section 5.2. To establish a balanced comparison, an equivalent number of tokens were randomly sampled from the model's vocabulary to serve as *normal* tokens. The seven tasks under consideration can be stratified into two primary types: Sentence-to-Sentence (S2S) and Sentence-to-Paragraph (S2P) tasks. For S2S tasks, sticky or normal tokens were inserted either at the start or end of a sentence. For S2P tasks, these tokens were inserted either at the beginning or the end of a paragraph. The quantity of tokens added was strategically set to constitute 10% of the original sentence or paragraph's token length.

**Results** Table 9 shows the results of our evaluation on 16 tasks of 7 categories. Compared with normal tokens, sticky tokens demonstrate higher destructiveness.

## G Explainability of Causes details

In this section, we attempt to investigate the underlying causes of the sticky token phenomenon. To systematically explain this phenomenon, we compare the intermediate results extracted from model layers and analyze the observed attention patterns and layer-wise divergence between sticky tokens and normal tokens.

**Setup** We experimented with the ST5-base model by creating a dataset of 1,000 sentences sampled from English Wikipedia. These sentences cover various topics to ensure generalizability. For analysis, we utilized the validated sticky tokens from Section 5.2. To establish a balanced comparison, an equivalent number of tokens were randomly sampled from the model's vocabulary to serve as normal tokens.

For each sentence, we generated two variants: 1) Sticky Token Variant: The original sentence inserted with a sticky tokens validated in Section 5.2 (e.g., </s>, lucrarca). 2) Normal token Variant: The original sentence inserted with a normal tokens randomly selected from the model's vocabulary.

We select a key feature to represent the model's internal state, i.e., *attention patterns*. The attention patterns capture the relative importance and relationships between tokens, providing insights on how the model synthesizes and modulates new representations within the attention head.

**Attention Pattern Disparity** Self-attention mechanisms in Transformer-based models dynamically allocate weights to tokens based on their contextual relevance.

Given an input sequence $X \in \mathbb{R}^{n \times d}$, where $n$ is the sequence length and $d$ is the embedding dimension, self-attention linearly projects $X$ into query, key, and value representations, i.e., $Q$, $K$, and $V$. The attention scores matrix $\mathcal{A}$ is then computed by taking the dot product between the query and key matrices, followed by a softmax normalization. The attention output is obtained by multiplying the attention scores with the value matrix.

---
[15]Evaluating Mistral-7B on the full MTEB benchmark requires over 40 hours using 8x A100 GPUs.

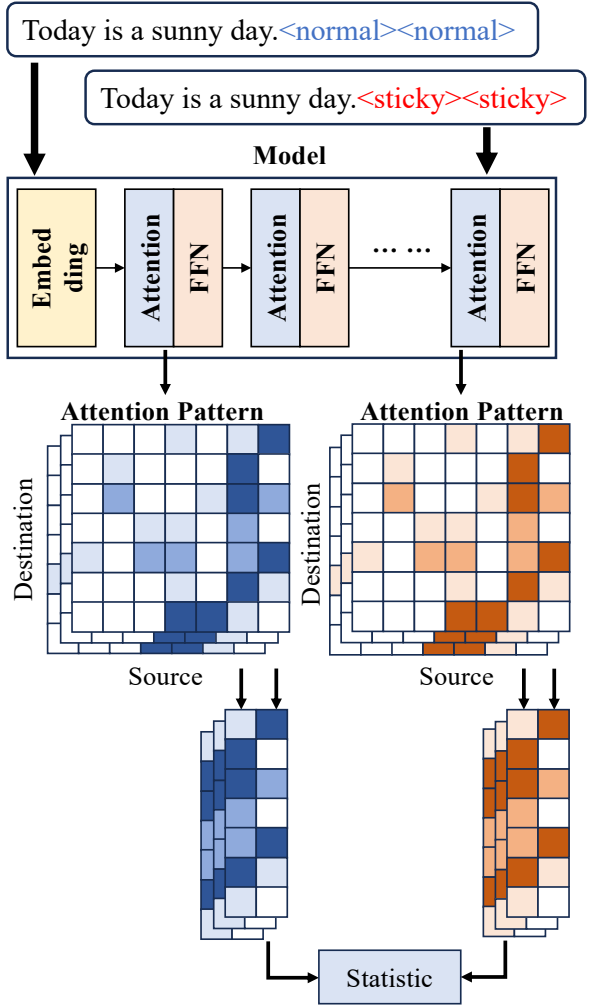

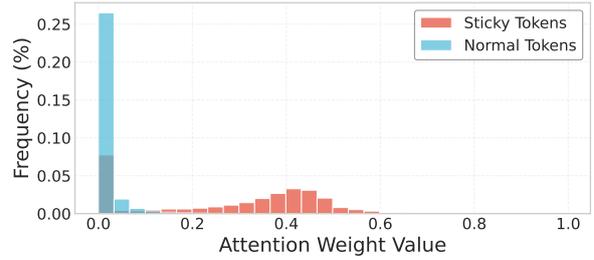

Figure 15: The example distribution of attention patterns. Sticky tokens (red) exhibit higher frequency in high-attention regions (>0.4) compared to normal tokens (blue).

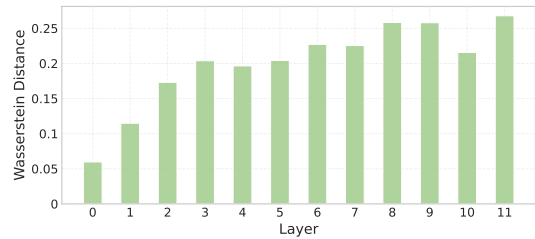

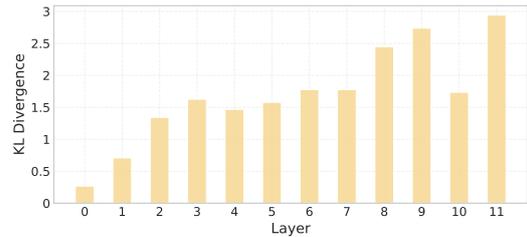

Figure 16: Wasserstein distance and KL divergence of the probability distributions between sticky tokens and normal tokens in different intermediate layers of ST5-base model.

Figure 14: A diagram of how to calculate the attention patterns of sticky and normal tokens.

$$A = \text{softmax}\left(\frac{QK^\top}{\sqrt{d}}\right)$$

$$\text{Attention}(Q, K, V) = A \cdot V$$

To analyze the behavior of Transformer-based models during sequence processing, we introduce the concept of attention patterns, which can be extracted from the corresponding column $A[:, n]$ of the attention scores matrix $A$. For a bidirectional encoder like ST5-base, the attention scores are computed across all tokens in the input sequence without masking.

As illustrated in Figure 14, for each sentence and attention head, we extract the values along the destination dimension of the attention score matrix at the position of the added token. Next, a comprehensive statistical analysis is performed to discern the patterns between sticky tokens and normal tokens.

Our analysis of attention scores reveals that sticky tokens exhibit distinct attention patterns compared to normal tokens. As shown in Figure 15, when sticky tokens are inserted to sentences, their attention weights in intermediate layers concentrate disproportionately in high-value ranges (e.g., weights > 0.4), whereas normal tokens follow a smoother, more Gaussian/Normal distribution. This suggests that sticky tokens dominate the model's focus and disrupt the balanced contextual representation of input texts.

**Layer-Wise Amplification of Anomalies** To illustrate the anomalies across different layers,

we employ the Wasserstein distance (Vaserstein, 1969) to quantify the differences in the outputs of intermediate layers generated by normal and sticky tokens. This approach helps uncover the variations in the model's internal mechanisms when processing these two types of tokens. In this study, a larger Wasserstein distance signifies a more significant divergence in distributions.

The Wasserstein distance (Vaserstein, 1969) between the attention patterns of sticky and normal tokens (Figure 16) further elucidates how anomalies propagate across layers.(We also plotted the graph of KL divergence, which is similar to the Wasserstein distance, as shown in Figure 16.) In early layers (1–6), the divergence remains moderate, indicating that shallow processing retains some robustness. However, from mid to late layers (6–12), the distance increases sharply, peaking at the final layers. This reflects a compounding effect: minor irregularities in early layers are progressively amplified as deeper layers integrate higher-order semantic features.

For text embedding models, the amplification disrupts the hierarchical abstraction of semantics. The anomalous intermediate results caused by sticky tokens are not uniformly distributed across all layers of the model but are concentrated and amplified in specific key layers.

## H Practical Implications and Mitigation

### H.1 Implication: Adversarial Attacks on LLM RAG Systems

A promising direction is leveraging sticky tokens for adversarial attacks in Retrieval-Augmented Generation (RAG) systems. RAG operates by:

- Retrieving documents semantically similar to a query using text embeddings.
- Generating answers conditioned on the retrieved documents.

Assuming that LLMs implicitly trust the embedding model's outputs, sticky tokens could be exploited to manipulate retrieval results. For instance, by injecting sticky tokens (e.g., lucrarea) into documents with toxic content, those documents may abnormally cluster near benign queries due to a *mean-pulling* effect. Consequently, toxic documents might dominate the retrieval results for otherwise innocuous queries (e.g., "how to improve mental health?"), potentially creating the risk of poisoning the LLM's output.

### H.2 Mitigation: Potential Strategies

We propose two mitigation strategies for addressing sticky tokens, and more strategies can be explored in future work.

**Tokenizer Sanitization** Most embedding models are fine-tuned from pre-trained foundation models (e.g., T5 (Raffel et al., 2020), BERT (Devlin et al., 2018)), inheriting their tokenizers. During fine-tuning, these tokenizers may include problematic tokens (e.g., *unused* tokens) that become sticky. A proactive measure could involve vocabulary pruning-removing tokens with abnormal frequencies (e.g., *unused* tokens, infrequent multilingual tokens, or non-ASCII characters) prior to fine-tuning the embedding model. However, the impact of adjusting the corresponding token embedding layer parameters due to modifications in the model's vocabulary remains to be explored in future work.

**Runtime Detection** For deployed models, a lightweight detector could flag input texts containing suspected sticky tokens (e.g., tokens with extreme frequency or abnormal positional distributions). Once detected, these tokens could be masked or re-embedded through context-aware recalibration.

## I Compute Statement

Most experiments presented in this paper were conducted on a computing cluster (PowerEdge XE9680 server) equipped with 8 NVIDIA A100 GPUs (80GB memory) running Ubuntu 22.04. We implement our framework in Python and use downloaded model checkpoints from Hugging Face. For all models, we employed 32-bit floating-point precision (fp32/float32) with standard default configurations.